\documentclass{article}

\usepackage{microtype}
\usepackage{graphicx}
\usepackage{subcaption}
\usepackage{booktabs} 

\usepackage{hyperref}

\usepackage[accepted]{eiml_icml2026}

\usepackage{amsmath}
\usepackage{amssymb}
\usepackage{mathtools}
\usepackage{amsthm}

\usepackage{multirow}
\usepackage{pifont}
\usepackage[table]{xcolor}
\usepackage{colortbl}
\usepackage{caption}
\newcommand{\std}[1]{\scriptsize{$\pm$ #1}}
\newcommand{\sota}[1]{\underline{#1}}
\usepackage{array}
\usepackage{adjustbox}

\usepackage[capitalize,noabbrev]{cleveref}

\theoremstyle{plain}

\theoremstyle{definition}

\theoremstyle{remark}

\icmltitlerunning{MAND: Modality-Aware Novelty Detection}

\begin{document}

\twocolumn[
  \icmltitle{MAND: Modality-Aware Novelty Detection \\for Open-World Egocentric Activity Recognition}



  \icmlsetsymbol{equal}{*}

  \begin{icmlauthorlist}
    \icmlauthor{Hyejeong Im}{equal,cau}
    \icmlauthor{Wonseon Lim}{equal,cau}
    \icmlauthor{Dae-Won Kim}{cau}
  \end{icmlauthorlist}

  \icmlaffiliation{cau}{Department of Computer Science and Engineering, Chung-Ang University, Seoul, Republic of Korea}
  \icmlcorrespondingauthor{Dae-Won Kim}{dwkim@cau.ac.kr}

  \icmlkeywords{Multimodal Learning, Novelty Detection, Open-World Continual Learning, Egocentric Activity Recognition}

  \vskip 0.3in
]













\printAffiliationsAndNotice{\icmlEqualContribution}

\begin{abstract}
Multimodal egocentric activity recognition integrates visual and inertial cues for robust first-person behavior understanding.
However, deploying such systems in open-world environments requires detecting novel activities while continuously learning from non-stationary data streams.
Existing methods rely on the main fused logits for novelty scoring, without fully exploiting the complementary evidence available from individual modalities.
Because these logits are often dominated by RGB, cues from other modalities, particularly IMU, remain underutilized, and this imbalance worsens as catastrophic forgetting accumulates.
To address this, we propose MAND, a modality-aware framework for multimodal egocentric open-world continual learning.
At inference, Modality-aware Adaptive Scoring (MoAS) adaptively adjusts modality contributions using sample-wise reliability and refines novelty scoring with deviation and disagreement penalties. 
During training, Modality-aware Representation Stabilization Training (MoRST) preserves the discriminative capacity of each modality across tasks through modality-specific heads and modality-wise logit distillation.
Experiments on a public multimodal egocentric benchmark show that MAND consistently improves novel activity detection and known-class accuracy while substantially reducing FPR95, indicating more reliable open-world recognition.
The source code is available at \href{https://github.com/HyeJeongIm/MAND}{github.com/HyeJeongIm/MAND}.
\end{abstract}

\begin{figure}[t!]
    \centering
    \includegraphics[width=\columnwidth]{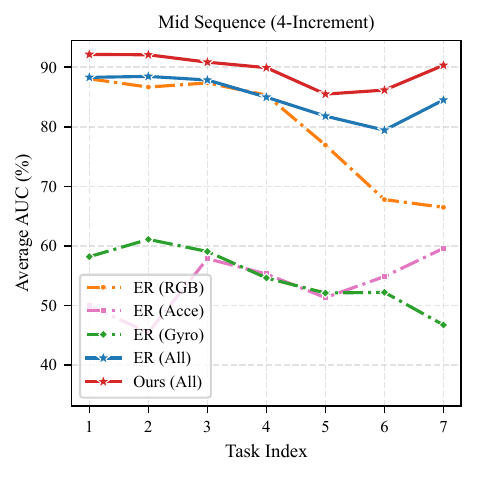}
    \caption{
        Task-wise average AUC for novel activity detection in the mid-sequence setting.
        ER (All) closely follows the RGB-only baseline, suggesting RGB-dominated main fused logits.
        MAND achieves consistently higher performance by better exploiting modality-specific evidence.
    }
    \label{fig:modality_dominance}
\end{figure}

\begin{figure*}[t!]
    \centering
    \includegraphics[width=\textwidth]{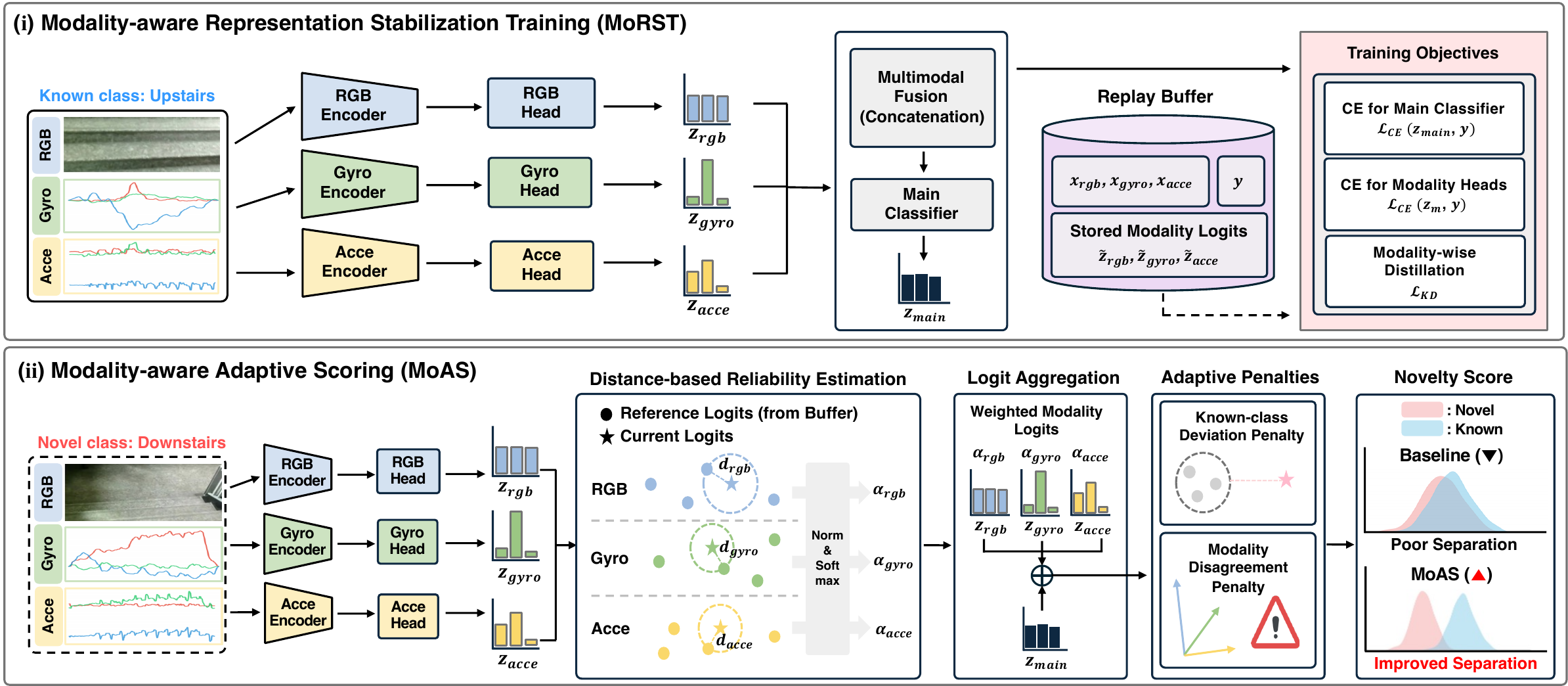}
    \vspace{-2em}
    \caption{
    Overview of the proposed MAND framework for multimodal egocentric open-world continual learning.
    \textbf{Top (MoRST, training):} modality-specific heads produce per-modality logits, while replay-based modality-wise logit distillation preserves modality-specific decision boundaries across continual tasks.
    \textbf{Bottom (MoAS, inference):} MoAS estimates distance-based modality reliability from replay-derived reference logits, aggregates weighted modality logits with the main fused logits, and applies deviation and disagreement penalties to produce the final novelty score. This leads to better separation between known and novel activities.
    }
    \label{fig:overview}
\end{figure*}

\section{Introduction}
Egocentric activity recognition has become a cornerstone of embodied AI, enabling assistive perception, AR/VR interaction, and life-logging applications~\cite{adrian2022egovis, grauman2022ego4d, engel2023projectaria}. 
Unlike third-person vision, egocentric perception provides first-person cues that reflect the wearer’s intent and motion~\cite{damen2022epickitchens}. 
Multimodal egocentric activity recognition (MMEA) integrates RGB with inertial measurement unit (IMU) signals, such as gyroscope and accelerometer streams, to provide complementary spatial and kinematic cues~\cite{gong2023mmg_ego4d}.
However, real-world deployment of MMEA requires detecting novel activities while continually learning from sequential task streams.

Recent works have extended MMEA to continual learning (CL) settings~\cite{xu2024mmea, he2024fgvir, cheng2024aid}, yet most assume a closed-world setting where only known classes appear at test time.
In real-world egocentric streams, unseen activities naturally appear over time, requiring models to detect novel activities while incrementally recognizing known ones~\cite{li2024prokt, kim2025owcl}.
This open-world continual learning (OWCL) setting further complicates MMEA, as the model must fuse heterogeneous modalities, distinguish novel activities from known ones, and continually adapt without catastrophic forgetting.

A recent effort, MONET~\cite{chee2025monet}, advances multimodal egocentric OWCL by incorporating modality-specific heads and pseudo-OOD-based dynamic thresholds into an online continual learning pipeline.
However, MONET relies solely on the main fused logits for novelty scoring, without leveraging modality-specific evidence from individual RGB and IMU logits. 
Because the main fused logits are primarily driven by RGB, the resulting novelty scores can underrepresent other modalities, particularly IMU, as illustrated in \cref{fig:modality_dominance}.
This imbalance becomes more pronounced over time, as IMU tends to be more susceptible to catastrophic forgetting, degrading its discriminability and reducing separability between known and novel activities.


\begin{table*}[t]
  \centering
  \caption{
        Novel activity detection results on UESTC-MMEA-CL under three class-incremental settings.
        Each CL method is evaluated with multiple novelty scoring functions.
        Results are mean $\pm$ std over 5 runs.
        Higher is better for $\mathbf{AUC}_{\mathrm{T}}$ ($\uparrow$), and lower is better for $\mathbf{FPR95}_{\mathrm{T}}$ ($\downarrow$).
        Bold indicates the best result, and underlining indicates the second-best result in each column.
        $\blacktriangledown$/$\vartriangle$ denotes statistically worse/better performance than MAND ($p<0.05$).
    }
  \label{tab:comp_novelty}

  \footnotesize
  \setlength{\tabcolsep}{4.2pt}
  \renewcommand{\arraystretch}{1.08}

  \begin{tabular}{c c c c c c c c}
  \toprule
  \multicolumn{2}{c}{\multirow{2}{*}{\textbf{Method}}}
  & \multicolumn{2}{c}{\textbf{Short Seq. (Inc: 8)}}
  & \multicolumn{2}{c}{\textbf{Mid Seq. (Inc: 4)}}
  & \multicolumn{2}{c}{\textbf{Long Seq. (Inc: 2)}} \\
  \cmidrule(lr){3-4} \cmidrule(lr){5-6} \cmidrule(lr){7-8}
  \multicolumn{2}{c}{}
  & $\mathbf{AUC}_{\mathrm{T}}$ ($\uparrow$) & $\mathbf{FPR95}_{\mathrm{T}}$ ($\downarrow$)
  & $\mathbf{AUC}_{\mathrm{T}}$ ($\uparrow$) & $\mathbf{FPR95}_{\mathrm{T}}$ ($\downarrow$)
  & $\mathbf{AUC}_{\mathrm{T}}$ ($\uparrow$) & $\mathbf{FPR95}_{\mathrm{T}}$ ($\downarrow$) \\
  \midrule

  \multirow{4}{*}{iCaRL}
  & MSP      & 85.92 \std{0.84}$\blacktriangledown$ & 57.86 \std{2.15}$\blacktriangledown$ & 84.51 \std{0.68}$\blacktriangledown$ & 63.83 \std{1.68}$\blacktriangledown$ & 81.33 \std{0.86}$\blacktriangledown$ & 66.60 \std{2.39}$\blacktriangledown$ \\
  & MaxLogit & 84.80 \std{0.95}$\blacktriangledown$ & 59.65 \std{4.85}$\blacktriangledown$ & 84.66 \std{1.03}$\blacktriangledown$ & 64.60 \std{1.24}$\blacktriangledown$ & \sota{82.18 \std{1.03}}$\blacktriangledown$ & 64.22 \std{2.89}$\blacktriangledown$ \\
  & Entropy  & 85.69 \std{0.85}$\blacktriangledown$ & 
  \sota{55.35 \std{1.68}}$\blacktriangledown$ & 85.02 \std{0.64}$\blacktriangledown$ & \sota{60.43 \std{1.36}}$\blacktriangledown$ & 81.75 \std{0.67}$\blacktriangledown$ & \sota{63.94 \std{2.36}}$\blacktriangledown$ \\
  & Energy   & 84.62 \std{0.97}$\blacktriangledown$ & 61.56 \std{5.28}$\blacktriangledown$ & 84.49 \std{1.05}$\blacktriangledown$ & 65.54 \std{1.54}$\blacktriangledown$ & 82.06 \std{1.05}$\blacktriangledown$ & 65.67 \std{2.68}$\blacktriangledown$ \\
  \midrule

  \multirow{4}{*}{ER}
  & MSP      & 84.04 \std{0.36}$\blacktriangledown$ & 59.63 \std{2.06}$\blacktriangledown$ & 82.61 \std{1.08}$\blacktriangledown$ & 66.13 \std{2.26}$\blacktriangledown$ & 80.02 \std{0.67}$\blacktriangledown$ & 69.35 \std{0.74}$\blacktriangledown$ \\
  & MaxLogit & \sota{86.37 \std{0.79}}$\blacktriangledown$ & 57.55 \std{5.18}$\blacktriangledown$ & \sota{85.06 \std{0.90}}$\blacktriangledown$ & 62.51 \std{3.03}$\blacktriangledown$ & 82.02 \std{1.01}$\blacktriangledown$ & 67.02 \std{1.32}$\blacktriangledown$ \\
  & Entropy  & 84.38 \std{0.30}$\blacktriangledown$ & 58.12 \std{2.28}$\blacktriangledown$ & 82.94 \std{1.09}$\blacktriangledown$ & 64.21 \std{1.68}$\blacktriangledown$ & 80.41 \std{0.65}$\blacktriangledown$ & 67.63 \std{0.84}$\blacktriangledown$ \\
  & Energy   & 86.22 \std{0.79}$\blacktriangledown$ & 58.87 \std{5.15}$\blacktriangledown$ & 84.90 \std{0.88}$\blacktriangledown$ & 63.73 \std{2.85}$\blacktriangledown$ & 81.79 \std{1.01}$\blacktriangledown$ & 67.90 \std{1.19}$\blacktriangledown$ \\
  \midrule

  \multirow{4}{*}{DER++}
  & MSP      & 82.05 \std{0.97}$\blacktriangledown$ & 64.80 \std{1.53}$\blacktriangledown$ & 81.68 \std{1.03}$\blacktriangledown$ & 70.24 \std{2.21}$\blacktriangledown$ & 78.43 \std{0.68}$\blacktriangledown$ & 73.33 \std{0.82}$\blacktriangledown$ \\
  & MaxLogit & 81.79 \std{1.00}$\blacktriangledown$ & 66.65 \std{4.02}$\blacktriangledown$ & 81.92 \std{1.48}$\blacktriangledown$ & 69.79 \std{2.08}$\blacktriangledown$ & 74.11 \std{1.31}$\blacktriangledown$ & 82.81 \std{1.73}$\blacktriangledown$ \\
  & Entropy  & 82.67 \std{1.01}$\blacktriangledown$ & 60.36 \std{2.52}$\blacktriangledown$ & 82.33 \std{1.09}$\blacktriangledown$ & 67.02 \std{2.53}$\blacktriangledown$ & 78.99 \std{0.71}$\blacktriangledown$ & 71.03 \std{1.00}$\blacktriangledown$ \\
  & Energy   & 81.48 \std{1.03}$\blacktriangledown$ & 68.07 \std{4.17}$\blacktriangledown$ & 81.65 \std{1.49}$\blacktriangledown$ & 71.45 \std{2.42}$\blacktriangledown$ & 73.56 \std{1.33}$\blacktriangledown$ & 84.63 \std{1.77}$\blacktriangledown$ \\
  \midrule

  \multirow{4}{*}{Foster}
  & MSP      & 83.81 \std{0.82}$\blacktriangledown$ & 60.41 \std{1.21}$\blacktriangledown$ & 80.85 \std{0.49}$\blacktriangledown$ & 68.26 \std{1.07}$\blacktriangledown$ & 76.57 \std{0.46}$\blacktriangledown$ & 72.20 \std{2.17}$\blacktriangledown$ \\
  & MaxLogit & 83.76 \std{1.02}$\blacktriangledown$ & 60.95 \std{2.50}$\blacktriangledown$ & 81.65 \std{0.40}$\blacktriangledown$ & 71.97 \std{0.70}$\blacktriangledown$ & 76.90 \std{0.42}$\blacktriangledown$ & 77.84 \std{2.01}$\blacktriangledown$ \\
  & Entropy  & 84.14 \std{0.76}$\blacktriangledown$ & 58.67 \std{0.52}$\blacktriangledown$ & 81.39 \std{0.49}$\blacktriangledown$ & 66.78 \std{1.03}$\blacktriangledown$ & 77.21 \std{0.45}$\blacktriangledown$ & 71.55 \std{1.41}$\blacktriangledown$ \\
  & Energy   & 83.61 \std{1.06}$\blacktriangledown$ & 62.70 \std{2.77}$\blacktriangledown$ & 81.44 \std{0.37}$\blacktriangledown$ & 73.26 \std{0.74}$\blacktriangledown$ & 76.34 \std{0.41}$\blacktriangledown$ & 79.78 \std{1.87}$\blacktriangledown$ \\
  \midrule

  \multirow{4}{*}{CMR-MFN}
  & MSP      & 70.86 \std{0.60}$\blacktriangledown$ & 79.23 \std{1.22}$\blacktriangledown$ & 67.58 \std{0.78}$\blacktriangledown$ & 85.69 \std{1.18}$\blacktriangledown$ & 58.70 \std{0.82}$\blacktriangledown$ & 89.08 \std{0.87}$\blacktriangledown$ \\
  & MaxLogit & 79.40 \std{0.33}$\blacktriangledown$ & 70.60 \std{1.39}$\blacktriangledown$ & 73.29 \std{0.63}$\blacktriangledown$ & 81.25 \std{1.77}$\blacktriangledown$ & 63.34 \std{0.85}$\blacktriangledown$ & 84.59 \std{1.40}$\blacktriangledown$ \\
  & Entropy  & 78.38 \std{0.35}$\blacktriangledown$ & 70.74 \std{0.75}$\blacktriangledown$ & 71.38 \std{0.58}$\blacktriangledown$ & 79.80 \std{1.35}$\blacktriangledown$ & 59.67 \std{1.09}$\blacktriangledown$ & 86.79 \std{1.92}$\blacktriangledown$ \\
  & Energy   & 79.43 \std{0.37}$\blacktriangledown$ & 70.94 \std{1.68}$\blacktriangledown$ & 73.20 \std{0.71}$\blacktriangledown$ & 81.44 \std{2.06}$\blacktriangledown$ & 63.60 \std{0.81}$\blacktriangledown$ & 83.75 \std{1.01}$\blacktriangledown$ \\
  \midrule

  \multirow{4}{*}{MONET}
  & MSP      & 84.54 \std{0.21}$\blacktriangledown$ & 59.40 \std{1.92}$\blacktriangledown$ & 83.39 \std{0.56}$\blacktriangledown$ & 66.72 \std{0.52}$\blacktriangledown$ & 80.07 \std{0.55}$\blacktriangledown$ & 68.90 \std{1.58}$\blacktriangledown$ \\
  & MaxLogit & 85.31 \std{0.55}$\blacktriangledown$ & 62.35 \std{3.93}$\blacktriangledown$ & 83.97 \std{0.75}$\blacktriangledown$ & 65.18 \std{1.64}$\blacktriangledown$ & 80.34 \std{0.90}$\blacktriangledown$ & 68.12 \std{2.44}$\blacktriangledown$ \\
  & Entropy  & 84.81 \std{0.26}$\blacktriangledown$ & 57.77 \std{1.30}$\blacktriangledown$ & 83.73 \std{0.61}$\blacktriangledown$ & 64.64 \std{0.73}$\blacktriangledown$ & 80.43 \std{0.57}$\blacktriangledown$ & 66.90 \std{1.78}$\blacktriangledown$ \\
  & Energy   & 85.06 \std{0.61}$\blacktriangledown$ & 64.11 \std{4.13}$\blacktriangledown$ & 83.81 \std{0.75}$\blacktriangledown$ & 66.38 \std{2.20}$\blacktriangledown$ & 80.17 \std{0.92}$\blacktriangledown$ & 69.06 \std{2.28}$\blacktriangledown$ \\
  \midrule

  \rowcolor{gray!10}
  \multicolumn{2}{c}{\textbf{MAND (Ours)}}
  & \textbf{89.43 \std{0.70}}\phantom{$\blacktriangledown$} & \textbf{44.93 \std{2.78}}\phantom{$\blacktriangledown$}
  & \textbf{89.59 \std{1.17}}\phantom{$\blacktriangledown$} & \textbf{47.62 \std{4.27}}\phantom{$\blacktriangledown$}
  & \textbf{86.02 \std{1.36}}\phantom{$\blacktriangledown$} & \textbf{55.31 \std{3.50}}\phantom{$\blacktriangledown$} \\
  \bottomrule
  \end{tabular}
\end{table*}


To address this, we propose MAND, a modality-aware framework for multimodal egocentric OWCL with two complementary components.
At inference, \textit{Modality-aware Adaptive Scoring} (MoAS) uses replay exemplars to estimate the reliability of each modality for a given sample with respect to known activities.
Based on this sample-wise reliability, MoAS adaptively adjusts the contribution of modality-specific logits and produces a novelty score that better reflects complementary modality evidence.
During training, \textit{Modality-aware Representation Stabilization Training} (MoRST) equips each modality encoder with a lightweight classification head and preserves per-modality decision boundaries across tasks via modality-wise logit distillation on replay exemplars.
By stabilizing these representations, MoRST provides more reliable modality-specific signals for novelty scoring at inference.

Extensive experiments on a public multimodal egocentric benchmark demonstrate the effectiveness of MAND.
MAND consistently outperforms competitive baselines across different sequence-length settings for both novel activity detection and known-class classification.
Notably, MAND achieves substantial reductions in FPR95, indicating more reliable rejection of novel activities near the known--novel boundary.
These consistent gains across settings highlight the importance of exploiting modality-specific evidence at both inference and training time for reliable multimodal egocentric OWCL.

The main contributions are summarized as follows:
\begin{itemize}
    \item We show that relying solely on the main fused logits for novelty scoring underutilizes modality-specific evidence due to modality dominance, which is further aggravated by modality-wise forgetting under incremental learning.
    \item We propose MoAS, an inference-time sample-adaptive scoring module that uses replay exemplars to estimate modality reliability and adjust modality-specific contributions for novelty scoring.
    \item We propose MoRST, a training-time module that preserves per-modality decision boundaries through modality-specific heads and logit distillation, maintaining stable modality-specific representations across tasks.
\end{itemize}

\section{Method}
\label{sec:method}

\subsection{Preliminary}
We study multimodal egocentric activity recognition~\cite{xu2024mmea} under an OWCL protocol~\cite{li2024prokt, kim2025owcl}. 
The learner is trained sequentially on a stream of class-incremental tasks $\{\mathcal{D}_t\}_{t=1}^{T}$, where each task introduces labeled samples from newly observed classes and expands the label space $\mathcal{Y}_0 \subset \mathcal{Y}_1 \subset \cdots \subset \mathcal{Y}_T$. 
Each training sample consists of three modalities, $x=\{x_{\text{rgb}},x_{\text{gyro}},x_{\text{acce}}\}$, with label $y\in\mathcal{Y}_t$. 
After learning task $t$, a test sample may belong either to a known class in $\mathcal{Y}_t$ or to a novel class outside $\mathcal{Y}_t$, requiring the model to jointly perform novelty detection and known-class classification under non-stationary multimodal streams. 

We adopt a standard multimodal encoder--fusion--classifier architecture~\cite{xu2024mmea,kazakos2019tbn}. 
Given the modality set $\mathcal{M}=\{\text{rgb},\text{gyro},\text{acce}\}$, each modality encoder $F_m(\cdot)$ produces an embedding $f_m = F_m(x_m)$, and a fusion module $G(\cdot)$ aggregates them into a fused representation $f_{\text{main}} = G(\{f_m\}_{m\in\mathcal{M}})$. 
The main classifier $C(\cdot)$ then outputs main fused logits $z_{\text{main}} = C(f_{\text{main}})$ for known-class classification.
In addition, we equip each modality encoder with a modality-specific head $H_m(\cdot)$ that outputs modality-specific logits $z_m = H_m(f_m)$, which are used for modality-aware novelty scoring and modality-wise representation stabilization. 

\begin{table*}[t!]
  \centering
  \caption{
    Scoring strategy comparison for novel activity detection under three class-incremental settings.
    Higher is better for $\mathbf{AUC}_{\mathrm{T}}$ ($\uparrow$), and lower is better for $\mathbf{FPR95}_{\mathrm{T}}$ ($\downarrow$).
    }
  \label{tab:tab_score_strategy}

  \begin{adjustbox}{width=\linewidth}
  \begin{tabular}{c c c c c c c}
  \toprule
  \multirow{2}{*}{\textbf{Scoring Strategy}}
  & \multicolumn{2}{c}{\textbf{Short Seq. (Inc: 8)}}
  & \multicolumn{2}{c}{\textbf{Mid Seq. (Inc: 4)}}
  & \multicolumn{2}{c}{\textbf{Long Seq. (Inc: 2)}} \\
  \cmidrule(lr){2-3} \cmidrule(lr){4-5} \cmidrule(lr){6-7}
  & $\mathbf{AUC}_{\mathrm{T}}$ ($\uparrow$) & $\mathbf{FPR95}_{\mathrm{T}}$ ($\downarrow$)
  & $\mathbf{AUC}_{\mathrm{T}}$ ($\uparrow$) & $\mathbf{FPR95}_{\mathrm{T}}$ ($\downarrow$)
  & $\mathbf{AUC}_{\mathrm{T}}$ ($\uparrow$) & $\mathbf{FPR95}_{\mathrm{T}}$ ($\downarrow$) \\
  \midrule

  Main Only ($\alpha_m=0$)
      & 86.82 \std{1.22} & 53.10 \std{4.37}
      & 87.01 \std{1.10} & 54.67 \std{3.83}
      & 83.50 \std{1.29} & 62.19 \std{1.71} \\
  Uniform ($\alpha_m=1/|\mathcal{M}|$)
      & 87.17 \std{1.09} & 50.00 \std{3.59}
      & 87.77 \std{1.15} & 51.13 \std{2.54}
      & 84.50 \std{1.07} & 59.02 \std{1.90} \\
  \rowcolor{gray!10}
  \textbf{MoAS (Ours)}
      & \textbf{89.43 \std{0.70}} & \textbf{44.93 \std{2.78}}
      & \textbf{89.59 \std{1.17}} & \textbf{47.62 \std{4.27}}
      & \textbf{86.02 \std{1.36}} & \textbf{55.31 \std{3.50}} \\
  \bottomrule
  \end{tabular}
  \end{adjustbox}
\end{table*}
\vspace{2em}
\begin{figure*}[t!]
  \centering
  \includegraphics[width=\textwidth]{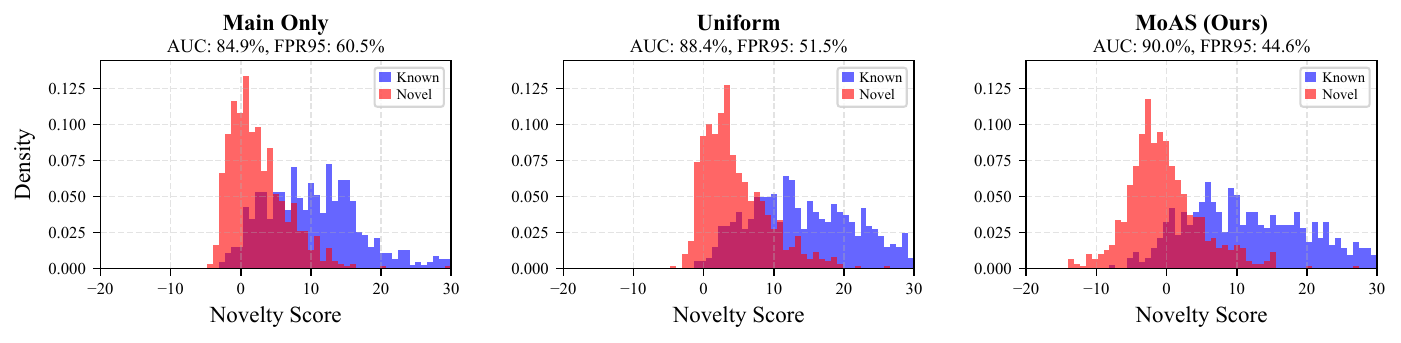}
  \caption{
  Effect of different scoring strategies on novelty score separability for Task 6 in the long-sequence (2-class incremental) setting.
  }
  \label{fig:scoring_strategy}
\end{figure*}

\subsection{Modality-Aware Novelty Detection and Training}
\label{sec:MAND}
We propose MAND, a modality-aware framework for multimodal egocentric OWCL, consisting of MoAS for inference-time novelty detection and MoRST for training-time representation stabilization.
As shown in \cref{fig:overview}, MoAS estimates sample-wise modality reliability from replay-based reference logits and adaptively combines modality-specific logits with the main fused logits for novelty scoring.
MoRST mitigates modality-wise representation drift through modality-specific heads and logit distillation.



\paragraph{Modality-aware Adaptive Scoring.}
\label{subsubsec:MoAS}

In open-world multimodal streams, the main fused logits $z_{\text{main}}$ can be dominated by a single high-confidence modality, suppressing complementary evidence and reducing known--novel separability.
To address this issue, we introduce Modality-aware Adaptive Scoring (MoAS).
MoAS estimates sample-wise modality reliability by comparing each modality logit with exemplar-derived logits in the current logit space.
This reliability is then used to adjust the contribution of modality-specific logits in the final logit aggregation.
Since adaptive weighting only changes relative modality contributions, MoAS further penalizes known-class deviation and modality disagreement during novelty scoring.

For modality $m$, let $\mathcal{R}$ denote the replay exemplar set stored in the replay buffer.
We construct a modality-specific reference logit set $\mathcal{L}_m$ by forwarding each exemplar in $\mathcal{R}$ through the current modality head.
We rebuild $\mathcal{L}_m$ after each task with the current model because the classifier head expands over tasks and modality representations can change during continual updates.
This ensures that replay exemplars and test samples are compared in the same current logit space.

Given a test sample $x$, we compute a normalized nearest-neighbor
distance between its modality logit $z_m(x)$ and the modality-specific
reference logit set $\mathcal{L}_m$:
\begin{equation}
\tilde{d}_m(x) =
\frac{
\min_{v \in \mathcal{L}_m} \|z_m(x)-v\|_2
-\mu_m
}{
\sigma_m
}.
\label{eq:normalized_dist}
\end{equation}
Here, $\mu_m$ and $\sigma_m$ denote the mean and standard deviation of
leave-one-out 1-NN distances within $\mathcal{L}_m$, respectively.
Smaller values of $\tilde{d}_m(x)$ indicate stronger consistency with
known-class reference logits.

We obtain modality weights by applying a temperature-scaled softmax to the negated normalized distances:
\begin{equation}
\alpha_m(x) = \frac{\exp(-\tilde{d}_m(x)/\tau)}{\sum_{j\in\mathcal{M}} \exp(-\tilde{d}_j(x)/\tau)}.
\label{eq:alpha}
\end{equation}
Here, $\tau$ is the temperature parameter that controls the sharpness of the modality-weight distribution.
These weights increase the contribution of modalities whose logits remain closer to the known-class reference set.
MoAS therefore adaptively determines how much each modality-specific logit contributes for each test sample.

To compute the novelty score, we construct the final logits by adding the modality-weighted logits to the main fused logits:
\begin{equation}
z_{\text{final},c}(x) = z_{\text{main},c}(x) + \sum_{m \in \mathcal{M}} \alpha_m(x)\, z_{m,c}(x).
\label{eq:final_logit}
\end{equation}

Although $\alpha_m(x)$ adjusts modality contributions, it does not directly penalize samples far from the known-class reference set because the softmax weights are relative and sum to one.
We therefore define a known-class deviation penalty:
\begin{equation}
P(x) = \frac{1}{|\mathcal{M}|}\sum_{m\in\mathcal{M}}\max\!\left(0,\tilde{d}_m(x)\right),
\label{eq:penalty}
\end{equation}
which lowers the score when modality logits deviate beyond the typical spread of reference logits.

We further define a modality disagreement penalty to capture cases that are not well explained by distance alone.
A near-novel sample may remain close to the known-class reference set and therefore have a small deviation penalty, while different modalities can still support conflicting class predictions.
To measure modality disagreement in a comparable space, we use predictive distributions rather than raw logits.
Let $q_m(x)=\mathrm{softmax}(z_m(x))$ and $q_{\text{main}}(x)=\mathrm{softmax}(z_{\text{main}}(x))$.
The mean predictive distribution over the main and modality-specific heads is
\begin{equation}
\bar{q}(x)=\frac{1}{|\mathcal{M}|+1}\!\left(q_{\text{main}}(x)+\sum_{m\in\mathcal{M}} q_m(x)\right).
\label{eq:qbar}
\end{equation}
The modality disagreement penalty is then defined as
\begin{equation}
\begin{aligned}
D_{\mathrm{KL}}(x)
= \frac{1}{|\mathcal{M}|+1}
\Bigg[
&\mathrm{KL}\!\left(q_{\text{main}}(x)\,\|\,\bar{q}(x)\right) \\
&+ \sum_{m\in\mathcal{M}}
\mathrm{KL}\!\left(q_m(x)\,\|\,\bar{q}(x)\right)
\Bigg].
\end{aligned}
\label{eq:dkl}
\end{equation}
A larger $D_{\mathrm{KL}}(x)$ indicates stronger modality disagreement.

The final novelty score is
\begin{equation}
s_{\mathrm{MoAS}}(x) =
\max_c z_{\mathrm{final},c}(x)
- \eta P(x)
- \gamma D_{\mathrm{KL}}(x).
\label{eq:score}
\end{equation}
The first term measures the strongest known-class evidence after reliability-weighted logit aggregation.
The second term penalizes deviation from the known-class reference set, and the third term penalizes modality disagreement.
The full MoAS inference procedure is summarized in Algorithm~\ref{alg:moas}.

\paragraph{Modality-aware Representation Stabilization Training.}
\label{subsubsec:MoRST}

The effectiveness of MoAS depends on the quality of per-modality evidence, which can degrade under continual updates. 
In particular, non-dominant modalities such as IMU are more susceptible to catastrophic forgetting than RGB-dominant fused predictions. 
This gradually reduces modality-specific discriminability and weakens modality-aware score fusion over the task stream. 
To preserve discriminative per-modality evidence throughout continual learning, we propose Modality-aware Representation Stabilization Training (MoRST).
During training, we jointly optimize the main classifier and modality-specific heads using the following objective:
\begin{equation}
\mathcal{L}_{\text{Sup}}
= \mathcal{L}_{\text{CE}}(z_{\text{main}}, y)
+ \lambda \cdot \frac{1}{|\mathcal{M}|} \sum_{m\in\mathcal{M}} \mathcal{L}_{\text{CE}}(z_m, y)
\label{eq:morst_cls}
\end{equation}
where $\mathcal{L}_{\text{CE}}$ denotes the cross-entropy loss and $\lambda$ controls the contribution of modality-specific supervision. 
This auxiliary supervision encourages each modality encoder to preserve class-discriminative structure in its own feature space, rather than relying solely on the dominant fused representation.

To further preserve modality-specific decision information across tasks, we store modality-specific logits together with replay exemplars in the buffer. 
For each exemplar $(\tilde{x},\tilde{y})\in\mathcal{B}$, we maintain stored modality-specific logits $\{\tilde{z}_m\}_{m\in\mathcal{M}}$, where $\tilde{z}_m$ denotes the previously recorded logits of modality $m$ for the same sample.
Unlike standard logit replay~\cite{buzzega2020der}, which matches a single fused output, MoRST penalizes deviations between each modality head and its own stored logits independently:
\begin{equation}
\mathcal{L}_{\text{KD}} = \sum_{m\in\mathcal{M}} \| \tilde{z}_m - z_m \|^2_2
\label{eq:morst_dist}
\end{equation}
This per-modality matching helps each modality encoder retain its own class-discriminative structure, yielding more stable modality-specific prediction signals that MoAS can leverage at inference. 
The final training objective is
\begin{equation}
\mathcal{L} = \mathcal{L}_{\text{Sup}} + \beta \cdot \mathcal{L}_{\text{KD}}
\label{eq:morst_total}
\end{equation}
with $\beta$ controlling the strength of modality-specific logit preservation.
The full MoRST training procedure is provided in Appendix~\ref{app:alg_morst}.


\section{Experiments}

\subsection{Datasets and Metrics}
We evaluate the proposed method on UESTC-MMEA-CL~\cite{xu2024mmea}, a multimodal egocentric benchmark consisting of RGB, gyroscope, and accelerometer streams over 32 activity classes. 
The experiments cover three class-incremental settings, with 8, 4, or 2 new activity classes introduced per task. We follow protocols for both novel activity detection and known-class classification.
For novelty detection~\cite{li2024prokt}, we report the area under the ROC curve ($\mathbf{AUC}_{\mathrm{T}}$) and the false positive rate at 95\% true positive rate ($\mathbf{FPR95}_{\mathrm{T}}$). 
For known-class classification~\cite{kim2025owcl}, we report the average accuracy ($\mathbf{ACC}_{\mathrm{T}}$) and forgetting ($\mathbf{FGT}_{\mathrm{T}}$) across all tasks.

\begin{table*}[t!]
  \centering
  \caption{
    Known-class classification results on UESTC-MMEA-CL under three class-incremental settings.
    Higher is better for $\mathbf{ACC}_{\mathrm{T}}$ ($\uparrow$), and lower is better for $\mathbf{FGT}_{\mathrm{T}}$ ($\downarrow$).
    Results are reported as mean $\pm$ std over 5 runs.
    $\blacktriangledown$/$\vartriangle$ denotes statistically worse/better performance than MAND ($p<0.05$).
  }
  \label{tab:comp_known}
  \begin{adjustbox}{width=\linewidth}
  \begin{tabular}{c c c c c c c}
  \toprule
  \multirow{2}{*}{\textbf{Method}}
  & \multicolumn{2}{c}{\textbf{Short Seq. (Inc: 8)}}
  & \multicolumn{2}{c}{\textbf{Mid Seq. (Inc: 4)}}
  & \multicolumn{2}{c}{\textbf{Long Seq. (Inc: 2)}} \\
  \cmidrule(lr){2-3} \cmidrule(lr){4-5} \cmidrule(lr){6-7}
  & $\mathbf{ACC}_{\mathrm{T}}$ ($\uparrow$) & $\mathbf{FGT}_{\mathrm{T}}$ ($\downarrow$)
  & $\mathbf{ACC}_{\mathrm{T}}$ ($\uparrow$) & $\mathbf{FGT}_{\mathrm{T}}$ ($\downarrow$)
  & $\mathbf{ACC}_{\mathrm{T}}$ ($\uparrow$) & $\mathbf{FGT}_{\mathrm{T}}$ ($\downarrow$) \\
  \midrule

  iCaRL
      & 85.96 \std{0.56}\phantom{$\blacktriangledown$} & 16.04 \std{0.93}\phantom{$\blacktriangledown$}
      & 82.90 \std{0.61}$\blacktriangledown$ & 17.48 \std{0.83}\phantom{$\blacktriangledown$}
      & 78.85 \std{0.91}$\blacktriangledown$ & 20.71 \std{1.11}$\blacktriangledown$ \\
  ER
      & 84.76 \std{1.29}\phantom{$\blacktriangledown$} & 17.56 \std{1.49}\phantom{$\blacktriangledown$}
      & 82.13 \std{1.66}$\blacktriangledown$ & 18.61 \std{1.80}$\blacktriangledown$
      & 77.92 \std{0.91}$\blacktriangledown$ & 21.85 \std{1.15}$\blacktriangledown$ \\
  DER++
      & 81.84 \std{1.21}$\blacktriangledown$ & 20.00 \std{1.52}$\blacktriangledown$
      & 79.16 \std{1.02}$\blacktriangledown$ & 20.92 \std{1.45}$\blacktriangledown$
      & 75.15 \std{1.59}$\blacktriangledown$ & 24.58 \std{1.59}$\blacktriangledown$ \\
  Foster
      & 79.48 \std{1.38}$\blacktriangledown$ & 24.70 \std{1.92}$\blacktriangledown$
      & 73.16 \std{1.23}$\blacktriangledown$ & 28.57 \std{1.53}$\blacktriangledown$
      & 67.66 \std{2.04}$\blacktriangledown$ & 32.79 \std{2.10}$\blacktriangledown$ \\
  CMR-MFN
      & 75.25 \std{1.40}$\blacktriangledown$ & \textbf{13.16 \std{1.31}}\phantom{$\blacktriangledown$}
      & 61.28 \std{1.41}$\blacktriangledown$ & \textbf{14.10 \std{1.11}}\phantom{$\blacktriangledown$}
      & 26.40 \std{2.07}$\blacktriangledown$ & 22.25 \std{1.91}$\blacktriangledown$ \\
  MONET
      & 85.26 \std{1.21}\phantom{$\blacktriangledown$} & 17.03 \std{1.33}\phantom{$\blacktriangledown$}
      & 82.29 \std{1.04}$\blacktriangledown$ & 18.42 \std{1.18}$\blacktriangledown$
      & 78.83 \std{1.01}$\blacktriangledown$ & 21.01 \std{1.02}$\blacktriangledown$ \\
  \midrule

  \rowcolor{gray!10}
  \textbf{MAND (Ours)}
      & \textbf{86.72 \std{1.03}}\phantom{$\blacktriangledown$} & 15.68 \std{1.14}\phantom{$\blacktriangledown$}
      & \textbf{85.11 \std{1.14}}\phantom{$\blacktriangledown$} & 15.39 \std{1.25}\phantom{$\blacktriangledown$}
      & \textbf{81.29 \std{1.45}}\phantom{$\blacktriangledown$} & \textbf{18.55 \std{1.36}}\phantom{$\blacktriangledown$} \\
  \bottomrule
  \end{tabular}
  \end{adjustbox}
\end{table*}

\subsection{Results}
We compare MAND with representative continual learning baselines, including replay-based methods and multimodal continual learning methods, paired with common post-hoc novelty scoring functions. 
These baselines primarily rely on the main fused logits for novelty scoring, whereas MAND explicitly incorporates modality-specific prediction signals from RGB and IMU streams.
Detailed descriptions of the compared methods are provided in Appendix~\ref{app:baselines}.

\paragraph{Novel Activity Detection Results.}
We compare MAND with all combinations of CL methods and novelty scoring functions under three class-incremental settings.
\Cref{tab:comp_novelty} reports AUC and FPR95 for each CL--scoring pair.
MAND achieves the best performance in all settings, improving AUC by 4.5\% and reducing FPR95 by 17.8\% on average over the best baseline for each metric and setting.
The FPR95 reduction is particularly important because it reflects fewer novel activities being incorrectly accepted as known at a strict operating point.
This suggests improved reliability near the known--novel boundary and more practical open-world deployment.
The task-wise AUC results in Appendix~\ref{app:novelty_results} further show that MAND maintains its gains throughout most of the task stream.

To analyze the effect of modality-aware scoring, \Cref{tab:tab_score_strategy} compares Main Only, Uniform, and MoAS.
Uniform outperforms Main Only, showing that modality-specific logits provide complementary evidence beyond the main fused logits.
MoAS further improves over Uniform, increasing AUC by 2.2\% and reducing FPR95 by 7.7\% on average.
This indicates that sample-wise reliability weighting separates known and novel activities more effectively than fixed aggregation.
\Cref{fig:scoring_strategy} shows that Main Only and Uniform still overlap between known and novel samples, whereas MoAS yields clearer separation.
Additional ablation studies on the proposed modules and MoAS components are provided in Appendix~\ref{app:ablation_moas}.

\paragraph{Known-Class Classification Results.}
\Cref{tab:comp_known} compares known-class classification performance across the three class-incremental settings. 
MAND achieves the best average accuracy across all sequence lengths, outperforming strong replay-based baselines such as iCaRL and ER, with the performance margin becoming larger as more tasks accumulate. 
MAND is particularly effective at mitigating forgetting in the long-sequence regime, where it achieves the lowest FGT among all compared methods.
This trend is further supported by the task-wise accuracy results in Appendix~\ref{app:taskwise_acc}, where MAND maintains higher accuracy throughout the task stream and degrades more slowly under the challenging 2-incremental setting.
These results suggest that preserving modality-specific representations improves both knowledge retention and known-class classification in multimodal egocentric continual learning.
Notably, the gains in known-class classification do not come at the expense of novelty detection, indicating that MAND improves stability and open-world adaptability simultaneously.

\section{Conclusion}
We introduced MAND, a modality-aware framework for open-world continual learning in multimodal egocentric activity recognition. MAND combines MoAS for sample-adaptive novelty scoring with MoRST for modality-wise representation stabilization. Across different sequence-length settings, MAND consistently improves both novel activity detection and known-class classification. These results show that preserving and exploiting modality-specific evidence is important for reliable multimodal egocentric OWCL.  
The current framework introduces additional memory costs from modality-specific heads and stored per-modality logits, and its effectiveness may depend on replay quality in longer task streams. Future work will explore more memory-efficient stabilization methods and broader multimodal settings with varying sensor availability and stronger distribution shifts. 


\bibliography{main}
\bibliographystyle{icml2026}

\newpage
\appendix
\onecolumn

\section{Related Work} 
\label{app:related_work}

\subsection{Multimodal Egocentric Activity Recognition} 
Human activity recognition has long benefited from fusing visual and inertial cues~\cite{chen2015utdmhad, ofli2013berkeleymhad}. More recently, large-scale egocentric datasets such as EPIC-KITCHENS~\cite{damen2022epickitchens}, Ego4D~\cite{grauman2022ego4d}, and ActionSense~\cite{delpreto2022actionsense} have enabled first-person activity understanding with synchronized RGB--IMU streams. UESTC-MMEA-CL~\cite{xu2024mmea} introduced the first multimodal egocentric benchmark for continual learning, together with a multimodal baseline evaluated using representative CL methods such as LwF~\cite{li2018lwf}, EWC~\cite{kirkpatrick2017ewc}, and iCaRL~\cite{rebuffi2017icarl}. CMR-MFN~\cite{wang2023cmrmfn} models evolving cross-modal correlations using an expandable architecture with confusion mixup regularization. AID~\cite{cheng2024aid} addresses modality imbalance by enhancing sensor features through vision--sensor attention. FGVIR~\cite{he2024fgvir} further improves long-term continual learning by reducing modality imbalance and promoting more generalizable multimodal representations. These studies highlight the importance of maintaining balanced and consistent modality representations in multimodal continual learning. Despite these advances, existing MMEA-CL methods remain confined to the closed-world assumption, and thus cannot detect novel activities that naturally arise in real-world egocentric streams. Unlike prior MMEA-CL methods, our work considers an open-world continual learning setting in which the model must jointly perform novel activity detection and known-class classification. 

\subsection{Open-World Continual Learning} 
Continual learning (CL) seeks to mitigate catastrophic forgetting through regularization, architectural expansion, or exemplar replay~\cite{mccloskey1989catastrophic,zenke2017si,rebuffi2017icarl}, but typically assumes a fixed label space that precludes detecting unseen categories. Open-world recognition~\cite{bendale2015openworld,wang2020novelty} highlights the need to jointly detect novel classes and retain prior knowledge, giving rise to \emph{Open-World Continual Learning} (OWCL)~\cite{liu2022aiautonomy}. \citet{joseph2021owod} identify unknown objects via contrastive clustering in an incremental detection setting. \citet{kim2025owcl} further establish novelty detection as a theoretical prerequisite for stable class-incremental learning, enabling self-initiated discovery and fast adaptation. MORE~\cite{kim2022more} and SHELS~\cite{gummadi2022shels} pursue representational disentanglement and feature exclusivity for novelty-aware CL, while Pro-KT~\cite{li2024prokt} transfers open-world knowledge through structured prototype evolution. All of these methods, however, target unimodal image streams. MONET~\cite{chee2025monet} is the first to extend OWCL to multimodal egocentric settings, combining online replay with pseudo-OOD thresholds and EMA-based distillation. However, its novelty scoring relies solely on the main fused logits, without leveraging complementary evidence from modality-specific logits, and it does not explicitly address modality-wise forgetting under sequential updates.

\section{Experimental Details}
\label{app:exp_details}

\subsection{Implementation}
\label{app:implementation}

For the backbone architecture, we use ImageNet-pretrained BN-Inception~\cite{ioffe2015bnincept,deng2009imgnet} for the RGB modality and DeepConvLSTM~\cite{ordonez2016deeplstm} for the IMU modalities to capture short- and long-term motion dynamics. All modalities are temporally aligned and fused using a Temporal Binding Network (TBN)~\cite{kazakos2019tbn}, and the fused features are fed into the classifier. 
Following UESTC-MMEA-CL~\cite{xu2024mmea}, we adopt modality-specific optimizers and hyperparameters. The RGB encoder is optimized with SGD, while the IMU encoders are optimized with RMSProp, both with an initial learning rate of 0.001. 
Each task is trained for 50 epochs, and the learning rate is decayed by a factor of 10 at epochs 10 and 20. The modality-specific loss weight $\lambda$ and distillation weight $\beta$ are set to 0.4 and 0.005, respectively, based on the validation set. 
To ensure fair comparison, we fix the replay buffer size to 320 for all replay-based methods. We evaluate three class-incremental settings with 8, 4, and 2 classes per task and report the mean and standard deviation over five random seeds. All experiments are implemented in PyTorch, conducted on a single NVIDIA RTX 3090 GPU, following the continual learning protocol of PyCIL~\cite{zhou2023pycil}.
In MoAS, the softmax temperature $\tau$, the deviation penalty weight $\eta$, and the cross-modal KL disagreement weight $\gamma$ are set to 3, 3, and 4, respectively. MoAS is used only for novelty detection, while known-class classification uses the main fused logits.

\subsection{Baselines}
\label{app:baselines}

We compare MAND against representative methods for both novelty scoring and continual known-class classification.
For novelty scoring, we consider widely used post-hoc methods, including MSP~\cite{hendrycks2016msp}, MaxLogit~\cite{hendrycks2022maxlogit}, Entropy~\cite{chan2021entropy}, and Energy~\cite{liu2020energy}. 
For continual known-class learning, we evaluate representative replay-based methods, including iCaRL~\cite{rebuffi2017icarl}, ER~\cite{chaudhry2019er}, DER++~\cite{buzzega2020der}, and Foster~\cite{wang2022foster}, as well as multimodal CL methods, including CMR-MFN~\cite{wang2023cmrmfn} and MONET~\cite{chee2025monet}. 
In Table~\ref{tab:comp_novelty}, each CL method is paired with each novelty scoring function.
Unlike the baseline combinations, MAND is evaluated as a unified framework that jointly performs continual learning and modality-aware novelty scoring.

\section{Algorithms}
\label{app:algorithms}

\subsection{Modality-aware Adaptive Scoring}
\label{app:alg_moas}

\begin{algorithm}[h!]
    \caption{Modality-aware Adaptive Scoring (MoAS)}
    \label{alg:moas}
    \footnotesize
    \begin{algorithmic}
        \STATE \textbf{Input:} Sample $x = \{x_m\}_{m\in\mathcal{M}}$, encoders $\{F_m\}_{m\in\mathcal{M}}$, modality-specific heads $\{H_m\}_{m\in\mathcal{M}}$, fusion $G$, classifier $C$, reference logit sets $\{\mathcal{L}_m\}_{m\in\mathcal{M}}$, distance statistics $\{(\mu_m,\sigma_m)\}_{m\in\mathcal{M}}$, hyperparameters $\tau,\eta,\gamma$
        \STATE \textbf{Output:} MoAS novelty score $s_{\text{MoAS}}(x)$
        
        \FOR{$m \in \mathcal{M}$}
          \STATE $f_m \gets F_m(x_m)$
          \STATE $z_m \gets H_m(f_m)$
          \STATE $\tilde{d}_m \gets \left(\min_{v \in \mathcal{L}_m} \|z_m - v\|_2 - \mu_m\right)/\sigma_m$
          \STATE $q_m \gets \mathrm{softmax}(z_m)$
        \ENDFOR
        
        \STATE $f_{\text{main}} \gets G(\{f_m\}_{m\in\mathcal{M}})$
        \STATE $z_{\text{main}} \gets C(f_{\text{main}})$
        \STATE $q_{\text{main}} \gets \mathrm{softmax}(z_{\text{main}})$
        
        \FOR{$m \in \mathcal{M}$}
          \STATE $\alpha_m \gets \exp(-\tilde{d}_m/\tau)\Big/\!\sum_{j\in\mathcal{M}}\exp(-\tilde{d}_j/\tau)$
        \ENDFOR        
        
        \STATE $z_{\text{final},c} \gets z_{\text{main},c} + \sum_{m\in\mathcal{M}} \alpha_m z_{m,c}$ for all $c$
        \STATE $P(x) \gets \frac{1}{|\mathcal{M}|}\sum_{m\in\mathcal{M}}\max(0,\tilde{d}_m)$
        \STATE $\bar{q} \gets \frac{1}{|\mathcal{M}|+1}\!\left(q_{\text{main}} + \sum_{m\in\mathcal{M}} q_m\right)$
        \STATE $D_{\mathrm{KL}}(x) \gets \frac{1}{|\mathcal{M}|+1}\!\left[\mathrm{KL}(q_{\text{main}}\|\bar{q}) + \sum_{m\in\mathcal{M}} \mathrm{KL}(q_m\|\bar{q})\right]$
        \STATE $s_{\text{MoAS}}(x) \gets \max_c z_{\text{final},c} - \eta P(x) - \gamma D_{\mathrm{KL}}(x)$
    \end{algorithmic}
\end{algorithm}

\subsection{Modality-aware Representation Stabilization Training} 
\label{app:alg_morst}

\begin{algorithm}[h!]
    \caption{Modality-aware Representation Stabilization Training (MoRST)}
    \label{alg:morst}
    \footnotesize
    \begin{algorithmic}
        \STATE \textbf{Input:} Training data $\mathcal{D}_t$, replay buffer $\mathcal{B}$, modalities $\mathcal{M}$, head loss weight $\lambda$, distillation weight $\beta$
        \STATE \textbf{Initialize:} Model parameters $\theta$
        
        \FOR{mini batch in $\mathcal{D}_t$}
          \STATE $(x, y) \gets \text{mini batch} \cup \mathcal{B}$
          \STATE $\{\tilde{z}_m\}_{m \in \mathcal{M}} \gets \mathcal{B}$
          \STATE $f_m \gets F_m(x_m)$ for all $m\in\mathcal{M}$
          \STATE $f_{\text{main}} \gets G(\{f_m\}_{m\in\mathcal{M}})$
          \STATE $z_{\text{main}} \gets C(f_{\text{main}})$
          \STATE $z_m \gets H_m(f_m)$ for all $m \in \mathcal{M}$
          \STATE $\mathcal{L}_{\text{Sup}}
          \gets \mathcal{L}_{\text{CE}}(z_{\text{main}}, y)
          + \lambda \cdot \frac{1}{|\mathcal{M}|} \sum_{m\in\mathcal{M}} \mathcal{L}_{\text{CE}}(z_m, y)$
          \STATE $\mathcal{L}_{\text{KD}}
          \gets \sum_{m\in\mathcal{M}}\| \tilde{z}_m - z_m \|^2_2$
          \STATE $\mathcal{L} \gets \mathcal{L}_{\text{Sup}} + \beta \cdot \mathcal{L}_{\text{KD}}$
          \STATE $\theta \gets \text{Optimizer}(\nabla \mathcal{L}, \theta)$
        \ENDFOR
    \end{algorithmic}
\end{algorithm}

\section{Results}
\label{app:results}

\subsection{Novel Activity Detection Results} 
\label{app:novelty_results}

\begin{figure*}[h!]
    \centering
    \includegraphics[width=0.9\textwidth]{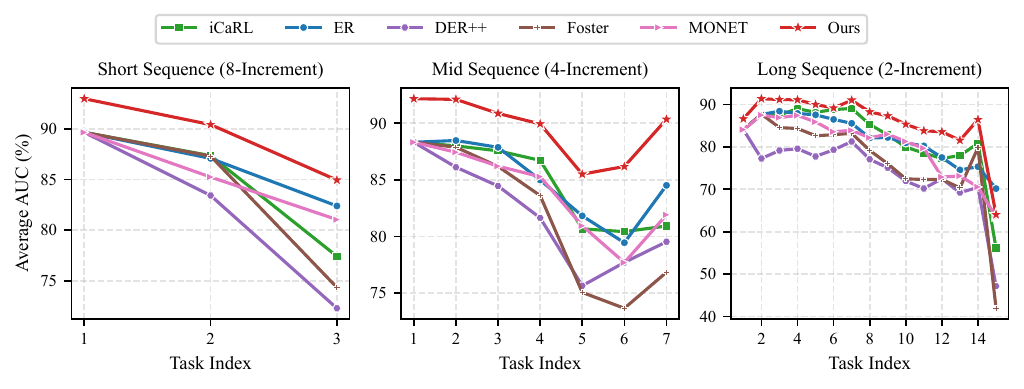}
    \caption{
    Task-wise novel activity detection performance.
    Average AUC over the task stream for the best-performing method pair in each class-incremental setting.
    }
    \label{fig:taskwise_auc_all}
\end{figure*}

\subsection{Known-Class Classification Results}
\label{app:taskwise_acc}    

\begin{figure*}[h!]
    \centering
    \includegraphics[width=\textwidth]{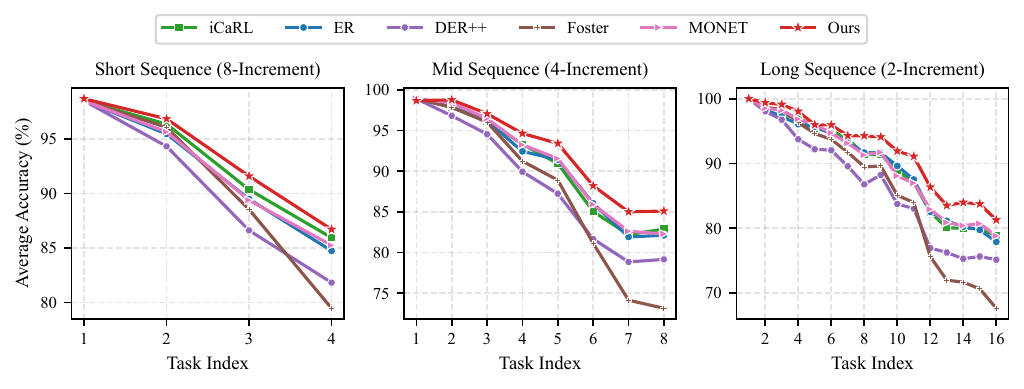}
    \caption{
    Task-wise known-class classification performance.
    Average accuracy over the task stream for continual learning methods under each class-incremental setting.
    }
    \label{fig:taskwise_acc_all}
\end{figure*}

\section{Ablation Study}
\label{app:ablation_moas}

\begin{table*}[t!]
  \centering
  \caption{
    Ablation results of the proposed modules for novel activity detection and known-class classification.
    Lower is better for $\mathbf{FPR95}_{\mathrm{T}}$ ($\downarrow$), and higher is better for $\mathbf{ACC}_{\mathrm{T}}$ ($\uparrow$).
  }
  \label{tab:tab_ablation}
  \begin{adjustbox}{width=0.8\linewidth}
  \begin{tabular}{c c c c c c c}
  \toprule
  \multirow{2}{*}{\textbf{Method}}
  & \multicolumn{2}{c}{\textbf{Short Seq. (Inc: 8)}}
  & \multicolumn{2}{c}{\textbf{Mid Seq. (Inc: 4)}}
  & \multicolumn{2}{c}{\textbf{Long Seq. (Inc: 2)}} \\
  \cmidrule(lr){2-3} \cmidrule(lr){4-5} \cmidrule(lr){6-7}
  & $\mathbf{FPR95}_{\mathrm{T}}$ ($\downarrow$) & $\mathbf{ACC}_{\mathrm{T}}$ ($\uparrow$)
  & $\mathbf{FPR95}_{\mathrm{T}}$ ($\downarrow$) & $\mathbf{ACC}_{\mathrm{T}}$ ($\uparrow$)
  & $\mathbf{FPR95}_{\mathrm{T}}$ ($\downarrow$) & $\mathbf{ACC}_{\mathrm{T}}$ ($\uparrow$) \\
  \midrule

  \rowcolor{gray!10}
  \textbf{MAND (Ours)}
      & \textbf{44.93 \std{2.78}} & \textbf{86.72 \std{1.03}}
      & \textbf{47.62 \std{4.27}} & \textbf{85.11 \std{1.14}}
      & \textbf{55.31 \std{3.50}} & \textbf{81.29 \std{1.45}} \\

   w/o MoRST
      & 45.66 \std{2.39} & 86.20 \std{0.93}
      & 51.14 \std{3.23} & 83.71 \std{1.55}
      & 57.14 \std{3.48} & 81.02 \std{1.48} \\

   w/o MoAS
      & 53.10 \std{4.37} & \textbf{86.72 \std{1.03}}
      & 54.67 \std{3.83} & \textbf{85.11 \std{1.14}}
      & 62.19 \std{1.71} & \textbf{81.29 \std{1.45}} \\

   w/o MoAS and MoRST
      & 57.55 \std{5.18} & 84.76 \std{1.29}
      & 62.51 \std{3.03} & 82.13 \std{1.66}
      & 67.02 \std{1.32} & 77.92 \std{0.91} \\

  \bottomrule
  \end{tabular}
  \end{adjustbox}
\end{table*}

\paragraph{Effect of the Proposed Modules.}
We conduct an ablation study to validate the contribution of each module in MAND.
\Cref{tab:tab_ablation} reports both novel activity detection and known-class classification performance under three class-incremental settings.
Removing MoAS leaves ACC nearly unchanged across all settings but consistently increases FPR95, indicating that its primary contribution lies in refining the novelty score rather than changing known-class classification.
By contrast, removing MoRST leads to lower ACC, especially in the mid-sequence setting, suggesting that MoRST is important for preserving reliable class-discriminative representations during continual learning.
Using either module alone improves over removing both, confirming that MoAS and MoRST contribute in complementary ways to balanced novelty detection and known-class classification.


\paragraph{Effect of MoAS Components.}
We further analyze the contribution of each component in MoAS in \cref{tab:tab_moas_ablation}.
The variant without both penalties retains only sample-adaptive reliability-weighted logit aggregation.
Although this already improves over using the main fused logits alone, it cannot directly penalize samples whose modality logits are collectively far from the known-class reference set.
The known-class deviation penalty addresses this limitation by lowering the score when modality logits exceed the typical reference-logit distance scale.
In contrast, the modality disagreement penalty captures a different failure case, where a near-novel sample remains close to the known-class reference set but different modalities support conflicting class predictions.
By penalizing disagreement among predictive distributions, this penalty complements the known-class deviation penalty.
Removing either penalty degrades performance, and removing both yields the weakest result.
This confirms that the distance and disagreement penalties capture complementary failure cases beyond reliability-weighted logit aggregation.

\begin{table*}[h!]
  \centering
  \caption{
    Ablation results of MoAS components for novel activity detection under three class-incremental settings.
    Higher is better for $\mathbf{AUC}_{\mathrm{T}}$ ($\uparrow$), and lower is better for $\mathbf{FPR95}_{\mathrm{T}}$ ($\downarrow$).
  }
  \label{tab:tab_moas_ablation}
  \begin{adjustbox}{width=0.85\linewidth}

  \begin{tabular}{c c c c c c c}
  \toprule
  \multirow{2}{*}{\textbf{Method}}
  & \multicolumn{2}{c}{\textbf{Short Seq. (Inc: 8)}}
  & \multicolumn{2}{c}{\textbf{Mid Seq. (Inc: 4)}}
  & \multicolumn{2}{c}{\textbf{Long Seq. (Inc: 2)}} \\
  \cmidrule(lr){2-3} \cmidrule(lr){4-5} \cmidrule(lr){6-7}
  & $\mathbf{AUC}_{\mathrm{T}}$ ($\uparrow$) & $\mathbf{FPR95}_{\mathrm{T}}$ ($\downarrow$)
  & $\mathbf{AUC}_{\mathrm{T}}$ ($\uparrow$) & $\mathbf{FPR95}_{\mathrm{T}}$ ($\downarrow$)
  & $\mathbf{AUC}_{\mathrm{T}}$ ($\uparrow$) & $\mathbf{FPR95}_{\mathrm{T}}$ ($\downarrow$) \\
  \midrule

  \rowcolor{gray!10}
  \textbf{MoAS}
      & \textbf{89.43 \std{0.70}} & \textbf{44.93 \std{2.78}}
      & \textbf{89.59 \std{1.17}} & \textbf{47.62 \std{4.27}}
      & \textbf{86.02 \std{1.36}} & \textbf{55.31 \std{3.50}} \\

    w/o KL Penalty
      & 88.84 \std{0.74} & 48.82 \std{3.00}
      & 89.19 \std{0.97} & 49.96 \std{2.88}
      & 85.73 \std{1.25} & 57.06 \std{2.55} \\

   w/o Distance Penalty
      & 88.30 \std{1.05} & 47.37 \std{3.12}
      & 88.98 \std{1.19} & 47.88 \std{2.90}
      & 85.50 \std{1.23} & 56.53 \std{3.13} \\

   w/o Distance and KL Penalties
      & 87.64 \std{1.12} & 51.00 \std{4.16}
      & 88.50 \std{1.01} & 50.27 \std{2.37}
      & 85.14 \std{1.14} & 58.47 \std{2.07} \\

  \bottomrule
  \end{tabular}
  \end{adjustbox}

\end{table*}

\end{document}